\documentclass[runningheads]{llncs}
\usepackage[T1]{fontenc}

\usepackage{graphicx}
\usepackage{listings}
\usepackage{dirtytalk}
\usepackage{times}
\usepackage{hyperref}
\usepackage{xcolor}
\usepackage{cite}
\usepackage{color}

\begin{document}
\title{Towards explainable decision support using hybrid neural models for logistic terminal automation}
%
%
\author{Riccardo D'Elia\inst{1}\orcidID{0009-0009-6172-4049} \and
Alberto Termine\inst{1}\orcidID{0000-0001-5993-0948} \and
Francesco Flammini\inst{1}\inst{2}\orcidID{0000-0002-2833-7196}}
\authorrunning{R. D'Elia et al.}

%
\institute{University of Applied Sciences and Arts of Southern Switzerland, Dalle Molle Institute for Artificial Intelligence, Lugano, Switzerland \email{name.surname@supsi.ch}
\and
University of Florence, Department of Mathematics and Computer Science Ulisse Dini, Florence, Italy 
\email{francesco.flammini@unifi.it}
 }


%
\maketitle              
\begin{abstract}
The integration of Deep Learning (DL) in System Dynamics (SD) modeling for transportation logistics offers significant advantages in scalability and predictive accuracy. However, these gains are often offset by the loss of explainability and causal reliability $-$ key requirements in critical decision-making systems. This paper presents a novel framework for interpretable-by-design neural system dynamics modeling that synergizes DL with techniques from Concept-Based Interpretability, Mechanistic Interpretability, and Causal Machine Learning. The proposed hybrid approach enables the construction of neural network models that operate on semantically meaningful and actionable variables, while retaining the causal grounding and transparency typical of traditional SD models. The framework is conceived to be applied to real-world case-studies from the EU-funded project AutoMoTIF\footnote{\url{https://automotif-project.eu}}, focusing on data-driven decision support, automation, and optimization of multimodal logistic terminals. We aim at showing how neuro-symbolic methods can bridge the gap between black-box predictive models and the need for critical decision support in complex dynamical environments within cyber-physical systems enabled by the industrial Internet-of-Things.

\keywords{artificial intelligence \and XAI \and IoT \and trustworthy autonomous systems \and transportation \and cognitive digital twins.}
\end{abstract}

\section{Introduction}
Intermodal terminals represent one of the most complex and critical environments within the modern logistics ecosystem. Functioning as key nodes where rail, road, and maritime systems intersect, these hubs must manage immense flows of containers, trailers, and vehicles in real time. With operations affected by volatile shipping schedules, fluctuating hinterland traffic, equipment availability, and infrastructure constraints, even minor disruptions can propagate through the system and lead to cascading delays and significant economic losses. In such dynamic settings, intelligent decision-support tools are essential $-$ not only to forecast and optimize system performance but also to provide explanations that are comprehensible to support trustworthy decisions and ease assessment by terminal operators, engineers, and regulators.

System Dynamics (SD) has long been employed as a foundational methodology to model and manage the complexity of logistics networks. SD models capture the evolution of systems over time through feedback loops, differential equations, and causal relationships grounded in domain expertise. Within terminal operations, SD has been applied to scenarios such as crane scheduling, gate throughput optimization, and yard congestion modeling, providing transparency and accountability through interpretable causal pathways~\cite{quaranta_review_2020,aschauer_modelling_2015,pirouzrahi_applying_2025}. These models have the advantage of semantic clarity, aligning with human mental models of causality and system behavior. However, traditional SD approaches suffer from key limitations: they rely heavily on expert-defined rules and assumptions, are constrained by oversimplified functional relationships, and become difficult to scale or calibrate as the number of interacting components and variables increases.

In recent years, Deep Learning (DL) has emerged as a powerful alternative, particularly for handling high-dimensional, nonlinear, and data-rich environments. DL has demonstrated considerable success in logistics for predictive tasks such as berth occupancy forecasting, crane movement duration estimation, or anomaly detection in terminal workflows~\cite{merkert_ai_2023, richey_artificial_2023, soumpenioti_ai_2023}. Leveraging historical operational data, DL models can autonomously extract latent patterns without explicit programming. However, this strength is also a weakness: DL architectures function as opaque \say{black-boxes}~\cite{sahin_unlocking_2025}, offering little to no insight into how decisions are made or what drives their predictions. More critically, these models often infer behavior based on statistical correlations rather than true causal mechanisms $-$ a problem referred to as \emph{causal reliability}~\cite{illari_causality_2024}. In contexts such as intermodal logistics, where safety, compliance, and operator accountability are paramount, the lack of interpretability\footnote{In this paper, we use the terms \emph{explainability} and \emph{interpretability} interchangeably to refer to the ability of AI systems to make their operations and decisions understandable to human stakeholders. While explainability typically emphasizes tailored explanations for users, and interpretability refers to the transparency of the model itself, both are treated here as complementary aspects of making AI behavior intelligible, trustworthy, and auditable~\cite{chazette_exploring_2021,mohseni_multidisciplinary_2021}.
} and causal transparency undermines the trustworthiness of purely data-driven models.

\subsubsection{Related Works.}
Within the paradigm of trustworthy artificial intelligence (AI), the field of Explainable AI (XAI) has arisen with the objective of understanding the internal rationale of deep learning models. Post-hoc XAI methods such as feature attribution, saliency maps, and surrogate modeling have been applied in logistics for use cases including predictive maintenance, risk analysis, and routing decisions~\cite{phan_systematic_2022,sharma_explainable_2024}. While these methods offer some visibility into input-output relationships, they remain fundamentally limited. First, most XAI techniques are applied after model training and do not alter the underlying architecture. Second, the explanations produced often lack semantic clarity or domain alignment, making them difficult for non-technical users to interpret or act upon. As noted in recent reviews of XAI in supply chain and logistics~\cite{elkhawaga_why_2024,kosasih_review_2024, yang_survey_2023}, this gap between technical insight and human interpretability remains a major barrier to adoption. The widespread use of DL methods in the modeling of dynamical systems calls for a paradigm shift to jointly address the challenges of interpretability and causal reliability. Although techniques exist for semantic interpretability, causal discovery, or mechanistic modeling in isolation, no one framework brings together (i) high-level, human-readable variables, (ii) robust interventional reasoning, and (iii) interpretable dynamic equations $-$ leaving a critical integration gap in complex, data-driven environments.
This paper advocates for precisely such an integration, bringing together advances from multiple fields to propose a methodological foundation for trustworthy optimization in logistics. In an increasingly digitalized world, such an approach for explainable decision support can be extended and generalized to additional classes of autonomous cyber-physical systems (CPS), in transportation and other domains~\cite{flammini_towards_2024}. From an infrastructural perspective, the approach is enabled by emerging technologies and paradigms, such as the Industrial Internet-of-Things (IIoT), interconnecting logistic terminal components, and Digital Twins (DTs), leveraging on the potential of cloud computing for model-based run-time monitoring, data analytics, and what-if projections~\cite{flammini_digital_2021,de_benedictis_digital_2023}.

\subsubsection{Novel Contribution.}

The approach is positioned as both a theoretical contribution and a practical roadmap for developing robust models of complex, high-risk environments such as intermodal terminals. Rather than focusing on the development and implementation of post-hoc explainability techniques, the proposed approach focuses on the construction of an \textbf{interpretable-by-design} neural system dynamics framework. This modeling framework combines the expressiveness of data-driven learning with the interpretability of symbolic reasoning. Rather than retrofitting explanations \emph{post-hoc}, this paradigm seeks to build models that are inherently explainable, causally reliable, and semantically aligned with real-world systems from the outset. A particularly compelling direction is the integration of Neuro-Symbolic AI, which combines the learning capabilities of neural networks with the reasoning structures of symbolic logic~\cite{bhuyan_neuro-symbolic_2024}. 

The new research direction proposed in this paper builds on these foundations to achieve a unified methodological approach for explainable decision making in intermodal logistics. At its core, this approach introduces an Interpretable Neural System Dynamics (INSD) pipeline designed to concurrently overcome the challenges of interpretability, scalability, and causal reliability in high-dimensional terminal operations optimization. This pipeline encompasses three main components. As a first step, Concept-Based Interpretability techniques are employed to extract semantically meaningful high-level variables from raw operational data. These \say{concepts} are aligned with domain-relevant metrics and behaviors, enabling human users to understand the factors driving system behavior~\cite{poeta_concept-based_2023}. Causal Machine Learning (CML) methods are then used to identify causal dependencies among the extracted concepts. Unlike conventional DL models that rely on statistical association, CML establishes cause–and–effect relationships, anchoring the learned model to the true generative processes of the system~\cite{kaddour_causal_2022, scholkopf_toward_2021}. Finally, mechanistically interpretable modeling techniques will be leveraged to infer a set of interpretable structural dynamic equations that govern the system’s behavior. These equations retain the mathematical transparency of SD while being learned from data, forming a hybrid structure that is both scalable and interpretable~\cite{bereska_mechanistic_2024}.

The resulting architecture is envisioned as the backbone of a new generation of Cognitive Digital Twins (CDTs) for intermodal terminals. These CDTs are not merely simulations of physical assets; rather, they are enriched with models that offer support for real-time monitoring, scenario analysis, and adaptive decision-making. Early research on CDTs has demonstrated potential in applications such as disruption recovery, predictive maintenance, and freight parking optimization~\cite{ashraf_disruption_2024, wasi_theoretical_2025, busse_towards_2021}, but most existing systems rely on black-box learning modules and lack semantic or causal integration.

The paper is organized as follows: Section~\ref{sec:background} reviews interpretable modeling concepts; Section~\ref{sec:pipeline} introduces the INSD pipeline; Section~\ref{sec:research_agenda} outlines research challenges; the final section concludes with key implications for decision support in intermodal logistics.

\section{Interpretable Modeling in Dynamic Logistics Systems}\label{sec:background}

The interpretability of models in logistics decision support systems is not just a matter of ease of use; it is an essential prerequisite for trust, accountability, and operational safety. In high-stakes environments like intermodal terminals, operators must be able to audit system behavior, trace recommendations back to intelligible causes, and simulate interventions under uncertainty. This section outlines three key components of an integrated interpretability framework for logistics optimization: System Dynamics, Explainable AI, and their hybridization into an appropriate modeling pipeline.

\subsection{System Dynamics: Modeling Causality in Complex Logistics Systems}

System Dynamics (SD) is a well-established modeling paradigm that represents the evolution of complex systems through interconnected feedback loops, stock-and-flow structures, and time delays. Originally developed for analyzing industrial processes and socio-economic systems, SD has proven especially effective in logistics, where resource flows and operational bottlenecks emerge from nonlinear interactions among interdependent components~\cite{quaranta_review_2020,aschauer_modelling_2015,pirouzrahi_applying_2025}.
At the heart of SD lies the \emph{causal loop diagram} (CLD), which provides a graphical language to describe how changes in one part of a system propagate to affect others. For example, an increase in crane utilization may reduce yard dwell time, thereby influencing gate throughput $-$ a feedback structure that can be formally captured in a CLD and translated into system equations. These diagrams enable both \emph{semantic interpretability} (clarity about what variables represent) and \emph{mechanistic interpretability} (how they interact). This explicit encoding of causality makes SD a reliable tool in modeling dynamic behaviors in terminal operations, from congestion buildup to equipment scheduling.
Despite these strengths, SD models are not without limitations. They are time-consuming to build, requiring deep domain knowledge and manual tuning. Moreover, they struggle to capture fine-grained nonlinearities and real-time adaptation, which are critical in modern, sensor-rich logistics environments. As terminal complexity grows, SD models may become insufficiently expressive or too coarse-grained, necessitating a complementary data-driven approach.

\subsection{Explainable Artificial Intelligence and the Interpretability Challenge}

\subsubsection{Opacity in Deep Learning for Logistics.}
In contrast to rule-based System Dynamics (SD) models, Deep Learning (DL) approaches provide unmatched scalability and predictive performance by learning complex input-output mappings directly from operational data. These models have been increasingly applied to a range of logistics tasks $-$ including berth occupancy prediction, crane sequencing, and dwell time estimation~\cite{awasthi_deep_2024,puli_predictive_2021,sobrie_capturing_2023}$-$ enabling real-time decision support across ports, intermodal terminals, and supply chains. However, this predictive power comes at a cost: DL models are notoriously \emph{opaque}, making it difficult for users to audit their reasoning or justify their outputs. This opacity is particularly problematic in logistics, where high-stakes decisions demand transparency, accountability, and regulatory compliance.

\subsubsection{Explainable AI Approaches.}
The field of \emph{Explainable Artificial Intelligence} (XAI) has emerged to overcome these black-box limitations by producing human-readable explanations for automated forecasts and recommendations~\cite{sokol_explainability_2022,kobayashi_explainable_2024}. Within critical logistics domains, XAI plays a growing role in enhancing human trust, operational safety, and system reliability~\cite{dolgui_towards_2021}. For instance, in the transport logistics domain, a hybrid explainable AI framework has been developed to analyze traffic accident data, enabling stakeholders to identify and interpret risk factors associated with different injury severity levels, thereby improving decision-making processes in road safety management~\cite{abdulrashid_explainable_2024}.

\subsubsection{Post-hoc vs.\ By-Design Explainability.}
Post-hoc XAI methods $-$ such as SHAP~\cite{lundberg_unified_2017} or LIME~\cite{ribeiro_why_2016} $-$ generate explanations after a model is fully trained. They offer the advantage of being model-agnostic and easy to apply, providing local, instance-level feature attributions. However, these explanations do not affect the underlying architecture; they may be misleading in the case of feature correlations or non-linear interactions, and do not guarantee that they reflect true causal pathways. By contrast, explainable-by-design approaches build transparency directly into the model’s structure~\cite{poeta_concept-based_2023}. Each component or parameter is explicitly tied to a human-understandable concept (e.g., \emph{yard congestion}, \emph{gate throughput}), and the architecture enforces modular, often sparse or additive forms that mirror causal loops and feedback mechanisms from system dynamics~\cite{muller_towards_2022}. Moreover, causal constraints can be embedded during training to ensure that learned relationships correspond to genuine cause-effect dependencies, thus supporting reliable interventional and counterfactual reasoning~\cite{scholkopf_toward_2021, pearl_causality_2009}. While such designs may impose limitations on expressive power or require careful adaptation to high-dimensional, time-dependent data, they deliver \emph{global} auditability and \emph{interventional} transparency $-$ properties essential for trust and accountability in safety-critical logistics applications.

\subsubsection{Semantic and Mechanistic Opacity.}
Despite this progress, interpretability remains an open and multi-faceted challenge. As outlined in recent literature~\cite{muller_towards_2022}, two major forms of opacity dominate discussions around XAI:

\begin{itemize}
    \item \emph{Semantic Opacity} refers to the difficulty of mapping a model’s learned internal representations to high-level, human-understandable concepts. Neural networks typically operate in high-dimensional feature spaces where latent variables do not correspond directly to domain-relevant categories. This discrepancy is particularly limiting in decision-making contexts, where well-founded explanations of operational terms are needed to gain insight into action.

    \item \emph{Mechanistic Opacity} involves the challenge of deciphering how various components of the model $-$ layers, neurons, weights $-$ interact to produce specific outputs~\cite{kastner_explaining_2024}. This issue is particularly significant in large-scale neural networks with millions or even billions of parameters, where computations are spread across numerous layers and involve several nonlinear transformations.
\end{itemize}
\noindent Most existing XAI methods fail to simultaneously address both semantic and mechanistic opacity, often focusing on one at the expense of the other~\cite{muller_towards_2022}.

\subsubsection{Causal Reliability.}
A third, often overlooked, dimension of the interpretability challenge is \emph{causal reliability}~\cite{illari_causality_2024, scholkopf_toward_2021}. This refers to a model’s ability to reflect the underlying causal mechanisms behind observed data, rather than merely exploiting statistical associations. DL models excel at identifying correlations but lack the structure to reason about cause and effect. This is a critical limitation in logistics operations, where decision-makers must evaluate the effects of critical management decisions (e.g., rerouting a shipment, modifying crane schedules). Without support for interventional and counterfactual reasoning~\cite{pearl_causality_2009}, such models fall short of enabling robust planning and adaptation to disruptions.

\subsubsection{Integrated Explainability Landscape.}
These three forms of opacity $-$ semantic, mechanistic, and causal $-$ are deeply interconnected. A model that is mechanistically transparent but semantically unintelligible offers limited utility; likewise, semantically aligned models without causal grounding cannot reliably support robust or fair decision-making. Current efforts in XAI and Causal Machine Learning (CML) have often developed in isolation, creating a fragmented research landscape~\cite{carloni_role_2023}. Their integration is crucial for modeling complex, dynamic systems like logistics networks, where interpretability must span across various dimensions. Recent surveys in logistics and supply chain research highlight the urgency of these challenges. For instance, Neuro-symbolic paradigms $-$ combining statistical learning with symbolic reasoning $-$ have been proposed to integrate domain knowledge into demand forecasting and inventory control~\cite{kosasih_review_2024}. However, most implementations still rely on post-hoc explanation techniques that fail to guarantee causal reliability. Even when integrated into routing and scheduling algorithms, DL systems struggle to explain the logic of complex, time-dependent decisions~\cite{wasi_theoretical_2025}. While these models may reduce costs, they often lack the transparency needed for intervention analysis and long-term planning.

This work responds to these limitations by proposing a cohesive framework that unifies the three domains of interpretability: concept-based (semantic) and mechanistic interpretability, and causal reliability. By integrating these perspectives within the modeling of dynamic logistic systems, the project seeks to bridge the gap between black-box predictive performance and human-aligned reasoning. This integrated approach will enable more reliable predictions and the ability to simulate interventions, analyze counterfactuals, and support human oversight in critical decision-making contexts.

\section{Toward Interpretable and Causally-Reliable Models}\label{sec:pipeline}

\begin{figure}[h]
  \centering
  \includegraphics[width=1\linewidth]{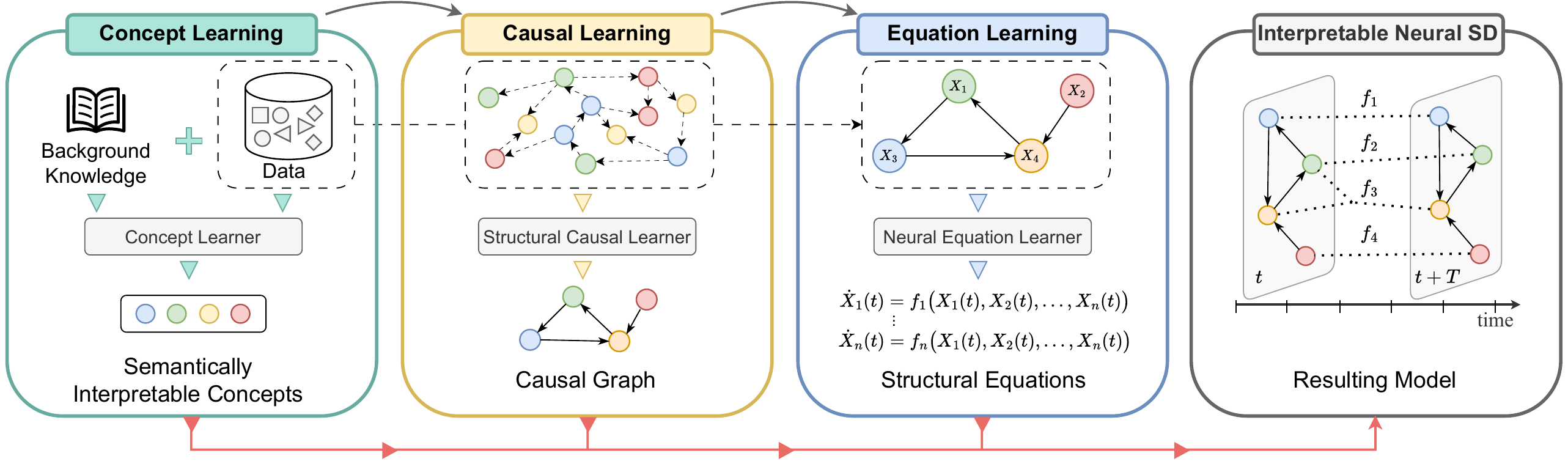}
    \caption{Overview of the INSD pipeline, from concept learning to causal and equation learning, ensuring interpretability and causal reliability in the resulting model.}
      \label{fig:INSD_pipeline}
\end{figure}

To unify the aforementioned threads, in this section we propose an \emph{Interpretable Neural System Dynamics (INSD)} pipeline $-$ a hybrid approach that integrates concept-based interpretability, causal discovery, and mechanistic modeling into a coherent framework (Figure~\ref{fig:INSD_pipeline}). The 3-step INSD pipeline addresses both semantic and mechanistic opacity while ensuring causal reliability across time-evolving systems:

\begin{enumerate}
    \item \textbf{Concept Learning}: Leveraging Concept-Based Interpretability (CBI), raw operational data is transformed into high-level variables (e.g., yard congestion, crane idleness) that align with human understanding~\cite{poeta_concept-based_2023}.
    \item \textbf{Causal Learning}: Causal Machine Learning and Causal Discovery methods are used to uncover dependencies among these concepts, producing a causal graph that encodes plausible cause-and-effect relations~\cite{kaddour_causal_2022}.
    \item \textbf{Equation Learning}: Finally, dynamic equations are learned from this causal graph using techniques from neuro-symbolic AI~\cite{brunton_discovering_2016,kramer_automated_2023}. This stage produces interpretable models capable of describing the temporal evolution of the system and enabling long-term intervention planning~\cite{bereska_mechanistic_2024}.
\end{enumerate}
Unlike standard neural networks, the resulting model can be queried, audited, and adapted in real time $-$ offering a balance of flexibility, performance, and explainability.

\subsection{Application of INSD Pipeline to Intermodal Terminals}
To illustrate how the proposed pipeline operates in practice, let us consider a digital twin simulation of an intermodal freight terminal. Here, a deep learning model forecasts recurring train delays during peak hours. Initial analyses show a correlation between these delays and increased truck traffic. However, without further interpretability, the underlying causes remain obscure. By integrating \emph{concept-based interpretability}, we can represent domain-relevant variables such as terminal workload, crane handling efficiency, and cross-modal wait times as explicit components in the model. This alignment allows predictions to be traced back to real-world operational concepts, ensuring that decision-makers understand not only what the model predicts but why.
To uncover whether truck congestion truly causes the train delays, \emph{causal reasoning} is applied. The model simulates counterfactual scenarios such as, \say{What if crane availability were increased during peak periods?} or \say{What if trucks were rerouted to a secondary access road?} This enables planners to test interventions and proactively identify leverage points within terminal operations.
Finally, the structural dynamic equations are used to track the flow of effects across the model's internal structure. The interpretable model reveals that the recommendation to reroute trucks stems from predicted access road bottlenecks and not from capacity shortfalls at the gates, as initially assumed. By disentangling the roles of modular components within the model, terminal operators can gain a structured, auditable understanding of how decisions are generated.
These forms of interpretability collectively enhance transparency, support trustworthy simulation-based decision-making, and empower human oversight in the operation of complex, dynamic logistics environments such as those addressed in the \emph{AutoMoTIF} project.

\subsection{Limitations of Existing XAI Approaches in Logistics}
Despite recent advances in Explainable AI for logistics, most existing approaches remain limited in their ability to support meaningful, actionable insight in dynamic operational settings. A key shortcoming is the lack of causal grounding. For example, methods like those proposed in~\cite{luo_explainable_2024} apply post-hoc explanation techniques to pre-trained deep models, providing insights into feature importance without distinguishing between mere statistical associations and true cause-and-effect relationships. These explanations are inherently descriptive rather than interventional, offering no guidance for counterfactual reasoning or scenario planning, essential tasks in high-risk domains such as intermodal logistics. Moreover, most existing models treat interpretability and predictive performance as a tradeoff rather than a design imperative. Additionally, explanations generated from latent neural activations often fail to align with semantically meaningful variables recognizable to domain experts, reinforcing the semantic opacity challenge discussed earlier. In short, current explainable systems either prioritize interpretability at the cost of accuracy or deliver high-performance predictions without ensuring that the outputs are causally valid or human-aligned. They fall short of enabling robust planning, adaptive decision-making, or regulatory accountability.

\subsection{INSD Pipeline Novelty and Impact}

The unified modeling framework advances the state of the art in both interpretability and causal reasoning for logistics decision-support systems. The proposed Interpretable Neural System Dynamics (INSD) pipeline is novel in three key dimensions:

\begin{itemize} \item \textbf{Interpretability-by-Design:} Rather than relying on post-hoc explanation, the INSD framework is designed from the ground up to be semantically and mechanistically interpretable. High-level concepts are learned explicitly from operational data and aligned with domain-relevant abstractions, facilitating intuitive understanding and traceability;
\item \textbf{Causally Reliable Modeling:} By integrating causal discovery and causal machine learning into the modeling pipeline, the INSD approach distinguishes correlation from causation, enabling robust interventional and counterfactual reasoning. This supports what-if analysis and decision-making under uncertainty;

\item \textbf{Structural Dynamic Equations Learning:} The final stage of the pipeline involves learning structural dynamic equations that retain the mechanistic interpretability of traditional SD models while leveraging the flexibility of neural architectures. 
\end{itemize}

\noindent This hybridization supports structural auditability, bridging the gap between symbolic modeling and deep learning. Moreover, the broader impact of this approach lies in its potential to serve as the core of next-generation Cognitive Digital Twins in logistics. By embedding causally reliable, interpretable models within digital replicas of intermodal terminals, the INSD framework can drive adaptive control, disruption response, and strategic planning with unprecedented transparency and trustworthiness. 

\section{Research Agenda and Open Challenges}\label{sec:research_agenda}

While the integration of System Dynamics and Deep Learning holds great promise for intermodal terminal decision support, realizing this vision demands advances across multiple fronts. Below, we present the key research challenges that must be addressed to build transparent, trustworthy, and operationally robust systems.

\subsubsection{Semantic Modeling of Operational Concepts.}
A foundational challenge lies in defining and extracting high-level, semantically rich concepts (e.g., yard congestion, crane idle time) from heterogeneous, multimodal data streams typical of modern terminals. Concept-based interpretability has shown promise in aligning model internals with human-understandable abstractions through techniques such as Concept Activation Vectors~\cite{kim_interpretability_2018} and hybrid causal-concept representation learning~\cite{goyal_explaining_2020}. However, most work focuses on static domains or language models; extending these methods to dynamic, streaming sensor data (RFID logs, camera feeds, IoT telemetry) demands novel architectures that can handle time-series invariance and contextual shifts across different terminal layouts.
Furthermore, learned concepts must remain robust under operational variability, such as equipment maintenance, procedural changes, or seasonal traffic fluctuations, while preserving alignment with operator mental models to facilitate rapid trust and adoption. This calls for adaptive concept learners that integrate domain ontologies and human-in-the-loop feedback, ensuring that representations not only capture statistical structure but also retain semantic fidelity in real-world settings~\cite{tao_semantic_2023}.

\subsubsection{Learning Causal Structures in Complex, Dynamic Environments.}
Intermodal operations involve feedback loops and delayed effects (e.g., a crane breakdown propagates to yard congestion hours later). Neural causal discovery methods have shown promise in uncovering static DAGs from observational data~\cite{scholkopf_toward_2021,kaddour_causal_2022}.  
Recent advances in spatio-temporal causal discovery, such as the SPACY framework, leverage variational inference to infer latent time-series causal graphs, offering a promising blueprint for terminal operations~\cite{wang_discovering_2024}. Yet, these methods typically assume dense observability and stationary processes, whereas terminals are characterized by partial data observation, ad-hoc interventions, and streaming constraints. Adapting causal discovery to this domain thus entails (i) robust handling of sensor noise and data gaps, (ii) support for cyclic feedback between subsystems, and (iii) online updating mechanisms that can incorporate new evidence without retraining from scratch. Embedding interventional calculus (do-calculus) will enable counterfactual reasoning critical for \say{what-if} scenario planning, such as quantifying the effect of deploying an extra crane during peak hours.

\subsubsection{Interpretable Modeling of System Dynamics.}
While System Dynamics (SD) has long offered white-box causal loop diagrams and differential equation models, it struggles with high-dimensional nonlinearity and real-time calibration. Mechanistic equation-learning methods $-$ such as Sparse Identification of Nonlinear Dynamics (SINDy) and physics-informed neural ODEs $-$ provide data-driven alternatives that can recover transparent governing equations from observations~\cite{brunton_discovering_2016}. Yet, applying these methods to the inherently discontinuous and multimodal dynamics of logistics terminals remains an open frontier. Key research tasks include devising modular equation-learning architectures that can isolate subsystem dynamics (e.g., crane cycles, truck arrival patterns) while preserving end-to-end predictive fidelity~\cite{wei_sparse_2022}. Moreover, embedding background domain knowledge, such as physical constraints and workflow rules, can regularize equation learners, yielding models that are both accurate and audit-friendly.

\subsubsection{Regulatory and Safety Considerations.}
As AI systems become increasingly embedded in critical infrastructure, ensuring their transparency, auditability, and compliance with emerging regulatory and ethical standards has become imperative. The complexity and high-risk nature of these domains demand not only effective but also trustworthy and accountable AI systems~\cite{bellogin_eu_2024}.
\emph{Transparency by design} is a foundational principle that must guide the development of decision-support models in these contexts. As emphasized by Felzmann et al., transparency should not be treated as a post-hoc feature but integrated throughout the system’s lifecycle, from data preprocessing to interface design~\cite{felzmann_towards_2020}. This ensures that models are intelligible and usable for domain operators, auditors, and policy stakeholders. \emph{Auditability} is crucial where decisions may trigger operational or safety consequences. Fernsel et al. propose a framework to assess AI systems' auditability, focusing on the availability of verifiable claims, structured evidential access, and tooling to support technical and procedural audits~\cite{fernsel_assessing_2024}. Their work outlines how auditability can be operationalized through traceable pipelines and documentation protocols.
To align these technical dimensions with ethical imperatives, \emph{ethics-based auditing} has emerged as a constructive practice for AI governance. Mokander and Floridi argue for continuous auditing processes that evaluate not only system outputs but also development practices, incentive structures, and the ethical alignment of goals~\cite{mokander_ethics-based_2021}. Such frameworks can function as a bridge between abstract ethical principles and deployable technical guidelines.
In parallel, the role of \emph{standardization} in governing AI transparency is gaining traction. H\"{o}gberg shows that transparency can be stabilized and institutionalized through technical standards that codify explanation practices and communication formats~\cite{hogberg_stabilizing_2024}. This is particularly relevant in logistics, where operational transparency must translate into consistent and interpretable outputs.
Finally, explainability remains central to both technical and organizational transparency. A recent analysis of explainability practices across AI ethics guidelines confirms that explainability is essential for understanding model behavior, building trust, and ensuring responsible deployment~\cite{balasubramaniam_transparency_2023}. The study underscores that effective implementation often requires multidisciplinary design teams that explicitly link explainability goals to stakeholder needs and system impacts.

\subsubsection{Towards an Integrated Evaluation Framework.}
To ensure that these principles are not only aspirational but also operational, there is a growing need for integrated evaluation frameworks that can assess AI systems along regulatory, ethical, and technical dimensions. For the INSD pipeline, this implies establishing metrics and traceability mechanisms for each layer of the architecture, from causal validity and semantic alignment to user comprehension and audit readiness. Such a framework would not only support internal quality assurance but also serve as a compliance interface for external audits and certification processes, bridging the gap between explainability, governance, and deployment in real-world logistics contexts. A comprehensive framework categorizing XAI design goals and corresponding evaluation methods has been proposed, facilitating iterative design and evaluation cycles in multidisciplinary teams~\cite{mohseni_multidisciplinary_2021}. This work underscores the importance of aligning XAI system design with specific user needs and evaluation strategies, which is crucial for developing transparent and accountable AI systems. Complementing this, a systematic literature review on ethics-based AI auditing highlights the necessity of integrating ethical principles $-$ such as fairness, transparency, and autonomy $-$ into AI system evaluations~\cite{laine_ethics-based_2024}. The review emphasizes the role of standardized methodologies in operationalizing these principles, ensuring that AI systems meet both ethical standards and stakeholder expectations. Incorporating these frameworks into the INSD pipeline can enhance its capacity to deliver AI decision-support systems that are not only functionally effective but also ethically sound and compliant with emerging regulatory standards.

\section*{Conclusion}\label{sec:conclusion}
This paper proposes a roadmap for combining semantic and mechanistic interpretability, causal reliability, and regulatory compliance into a cohesive framework within reference industrial applications. By charting a research agenda across concept identification, causal discovery, and equation learning, we shed light on the technical and organizational requirements for the operationalization of the Interpretable Neural System Dynamics (INSD) pipeline. In doing so, we envision a new generation of Cognitive Digital Twins for intermodal terminals: digital replications enabled by emerging CPS and IIoT technologies that not only forecast and optimize, but also reveal their reasoning in terms that operators can understand and regulators can audit~\cite{le_digital_2024,flammini_digital_2021,de_benedictis_digital_2023}. Beyond logistics, this approach exemplifies a broader paradigm shift toward interpretability-by-design in AI systems deployed in critical infrastructures, where trust, safety, and accountability are essential~\cite{flammini_towards_2024}. Realizing this vision will demand concerted, multidisciplinary collaboration among AI researchers, system dynamics experts, logistics practitioners, and policymakers. Rewards include enhanced resilience, informed decision-making, and societal trust, justifying this ambitious undertaking. The INSD framework offers not just a path forward for logistics but a template for embedding transparency and causality at the heart of AI-driven decision support across domains.

\begin{credits}
\subsubsection{\ackname} This work was partly supported by the Swiss State Secretariat for Education, Research and Innovation (SERI) under contract no. 24.00184 (AutoMoTIF project). The project has been selected within the EU Horizon Europe programme under grant agreement no. 101147693. Views and opinions expressed are however those of the authors only and do not necessarily reflect those of the funding agencies, which cannot be held responsible for them.
\end{credits}
%
%
%
%

\bibliographystyle{splncs04}
\bibliography{references} 

\end{document}